\documentclass[11pt,a4paper]{article}
\usepackage[hyperref]{emnlp-ijcnlp-2019}
\usepackage{times}
\usepackage{latexsym}
\usepackage{amsmath}
\usepackage{url}
\usepackage{multirow}
\usepackage{makecell}
\usepackage{array}
\usepackage{hhline}
\usepackage{booktabs}
\usepackage{algorithmic}
\usepackage{setspace}
\usepackage{graphicx}
\usepackage{stackengine}
\usepackage{xcolor}
\usepackage{soul}
\usepackage{bm}
\newcolumntype{L}[1]{>{\raggedright\let\newline\\\arraybackslash\hspace{0pt}}m{#1}}
\newcolumntype{C}[1]{>{\centering\let\newline\\\arraybackslash\hspace{0pt}}m{#1}}
\newcolumntype{R}[1]{>{\raggedleft\let\newline\\\arraybackslash\hspace{0pt}}m{#1}}
\newcommand{\hlc}[2][yellow]{{\colorlet{foo}{#1}\sethlcolor{foo}\hl{#2}}
}
\usepackage{tabularx} 

\aclfinalcopy 

\newcommand{\todo}[1]{} 
\newcommand{\modelname}[0]{\textsc{ProfileREG}}

\title{Referring Expression Generation Using Entity Profiles}

\author{Meng Cao\qquad Jackie Chi Kit Cheung \\ \\
    School of Computer Science, McGill University, Montreal, QC, Canada \\
    MILA, Montreal, QC, Canada \\ \\
    {\{\tt meng.cao@mail, jcheung@cs\}.mcgill.ca}}

\begin{document}
\maketitle
\begin{abstract}
Referring Expression Generation (REG) is the task of generating contextually appropriate references to entities. A limitation of existing REG systems is that they rely on entity-specific supervised training, which means that they cannot handle entities not seen during training. In this study, we address this in two ways. First, we propose task setups in which we specifically test a REG system's ability to generalize to entities not seen during training. Second, we propose a profile-based deep neural network model, \modelname{}, which encodes both the local context and an external profile of the entity to generate reference realizations. Our model generates tokens by learning to choose between generating pronouns, generating from a fixed vocabulary, or copying a word from the profile. We evaluate our model on three different splits of the WebNLG dataset, and show that it outperforms competitive baselines in all settings according to automatic and human evaluations.

\end{abstract}

\section{Introduction}

Entities can be expressed by various types of linguistic expressions, including by their names (e.g., \textit{Barrack Obama}), a definite description (e.g., \textit{the former president of the United States}), a pronoun (e.g., \textit{he}, \textit{him}, \textit{his}), or a demonstrative (e.g., \textit{that person}). Many factors play a role in determining what type of expression is appropriate in a particular context \cite{henschel2000pronominalization}, including information status, familiarity with the entity, and referential clarity. In this study, we aim to design a model that can generate appropriate referring expressions of entities in an extended passage. Such a system is useful in a variety of natural language generation settings, from dialogue systems to automatic summarization applications \cite{reiter2000building, krahmer2012computational}. 

Referring expression generation (REG) can be broken into two steps. The first is to decide the \emph{form} of referring expression; that is, what type of reference should be used (e.g., a proper name). The second is to determine the \emph{content} of the referring expression (e.g., \emph{Ernest Hemingway}). 

Many computational approaches, both rule-based and machine-learning-based, have been proposed for REG. Rule-based models use pre-defined heuristics and algorithms to determine referential form \cite{reiter2000building, henschel2000pronominalization, callaway2002pronominalization}. Machine learning-based approaches require training on samples to predict referring expressions \cite{nenkova2003references, greenbacker2009feature, ferreira2016towards, ferreira2017generating}, often as a classification task. 

The common limitation of previous REG systems is that they are incapable of generating referring expressions for new, unseen entities. Previous REG setups have tended to focus on form selection rather than content generation, or else they provide a static list of attributes or realization options for models to select from \citep{belz2010generating,gatt2008attribute}. The recent NeuralREG model generates referring expressions in an integrated, end-to-end setup, but it requires seeing instances of the entity being referred to in the training set \citep{ferreira2018neuralreg}.

In this work, we address this problem by proposing new REG task setups which test for REG systems' ability to handle new entities at test time. We also develop a REG system which can handle entities not seen during training using external knowledge about them, generated using extracts of their Wikipedia page.

From a practical perspective, it is reasonable to assume that such an entity profile exists for common entities such as popular locations and celebrities, as well as for targeted entities of interest that should be handled by an in-domain NLG system. We make the minimal assumption that the profile contains just a few sentences about the entity, so that these profiles can easily be written by non-experts or be automatically extracted.

Our model, \modelname{}, uses a learned switch variable to decide whether to generate a token from a fixed vocabulary, generate a pronoun, or use information from the profile in order to generate appropriate referring expressions in context. We evaluate our model on the WebNLG corpus \cite{gardent2017creating}. Experimental results show that our model is capable of handling unseen entities that prior work simply cannot handle in our new evaluation setups, while also outperforming them in the original setting.\footnote{https://github.com/mcao610/ProfileREG}

Our contributions are as follows. First, we address an important limitation in prior REG studies by creating new test setups that evaluate neural REG models specifically on entities that are not seen in the training set. Second, we propose a new REG model, \modelname{}, which outperforms existing REG models in all tested settings according to automatic and human evaluation measures.

\section{Related Work} \label{related_work}
Previous work in REG can be divided into two groups: rule-based and machine learning-based. 

Rule-based approaches use pre-defined rules or algorithms to determine the form or content of generated referring expressions. For example, \citet{reiter2000building} proposed a straightforward heuristic for REG: a proper name should be used in the first mention of an entity and a pronominal form should be used for the subsequent references if there is no mention of any other entity of the same gender between the reference and its antecedent. \citet{henschel2000pronominalization} presented an algorithm for the pronominalization of third person discourse entities. 

Machine learning-based approaches predict the form or content of referring expressions using features extracted from the context. Commonly used context features include syntactic position, referential status (initial or a subsequent mention) and distance. For instance, \citet{nenkova2003references} built a Markov chain model to capture how the subsequent referring expressions are conditioned by earlier mentions. \citet{frank2012predicting} proposed a rational speaker-listener model based on the assumptions that speakers attempt to be informative and that listeners use Bayesian inference to recover the speakers' intended referents. \citet{orita2015discourse} extended the previous model by introducing the referent's discourse salience and a cost term. \citet{ferreira2016towards} proposed a naive Bayes model to predict the probability of a particular referential form given a set of feature values, including syntactic position, referential status and recency of references. Similarly, \citet{ferreira2017generating} used a naive Bayes model based on syntactic position and referential status to predict the form of a proper name reference.

In recent years, deep neural networks have achieved great success in a variety of NLP applications (e.g., machine translation \cite{bahdanau2014neural} and automatic summarization \cite{rush-etal-2015-neural, see-etal-2017-get}). To the best of our knowledge, there have only been two models that use deep neural networks for REG. The first is by \citet{ferreira2016towards}, who use recurrent neural networks (RNNs) to predict the form of referring expressions. They use RNNs to encode a sequence of discourse features and apply a softmax layer to generate referential form distribution. 

The other is the NeuralREG model proposed by \citet{ferreira2018neuralreg}, an end-to-end system that predicts both the form and content of the referring expression. NeuralREG has an encoder-decoder structure, using LSTM units to encode contextual information and to decode referring expressions. Since each entity is represented as an embedding vector in the model, it cannot handle new entities that are outside the training set.

Generating entity descriptions based on source data has also been explored in the context of concept-to-text generation \citet{lebret-etal-2016-neural}. We focus here on generating referring expressions from unstructured source data, which may be more readily available in some settings.
\section{Data}
\label{section:data} 

The WebNLG corpus \cite{gardent2017creating} was initially introduced in the WebNLG challenge 2017 \cite{gardent2017webnlg}. The WebNLG dataset is made up of (data, text) pairs where the data is a set of triples that consists of entities and their relationships and the text is a verbalisation of these triples. In this challenge, the participants are asked to map the triple data to text.

In this study, we use a delexicalized version of WebNLG corpus introduced by \citet{ferreira2018neuralreg}. The authors manually extracted each referring expression in the text and kept only the ones referring to Wikipedia entities. As a result, each sample of the delexicalized dataset consists of a Wikipedia ID, a true-cased tokenized referring expression (extracted from the text) and a lowercased, tokenized discourse context preceding and following the target reference (referred as the pre- and post-context). Table~\ref{table:webnlg} shows one sample from the corpus. 
The dataset consists of 78,901 referring expressions for 1,501 Wikipedia entities. Among the expressions, 71.4\% (56,321) are proper names, 5.6\% (4,467) are pronouns, 22.6\% (17,795) are descriptions and 0.4\% (318) are demonstratives.

\begin{table}[t]
\small
\renewcommand{\arraystretch}{1.2}
\setlength\tabcolsep{2.5pt}
\centering
\begin{tabular}{|p{7.3cm}|}
  \hline
  {\bf Triples}: \\
    Elliot\_See $ \mid $ almaMater $ \mid $ University\_of\_Texas\_at\_Austin \\
    Elliot\_See $ \mid $ deathPlace $ \mid $ St.\_Louis \\
    Elliot\_See $ \mid $ birthPlace $ \mid $ Dallas \\
    ... \\
  {\bf Text}: Elliot See was born on July 23 , 1927 in Dallas . \textit{He} attended the U of Texas at Austin which is part of the U of Texas system . See was a test pilot . He died on Feb 28 , 1966 in St. Louis . \\
  \hline
  {\bf Wikipedia ID}: elliot\_see \\ 
  {\bf Referring expression}: He \\
  {\bf Delexicalized text}: elliot\_see was born on 1927-07-23 in dallas . elliot\_see attended the university\_of\_texas\_at\_austin which is part of university\_of\_texas\_system . elliot\_see was a test pilot . elliot\_see died on 1966-02-28 in st.\_louis . \\ 
  {\bf Pre-context}: elliot see was born on 1927-07-23 in dallas . \\ 
  {\bf Post-context}: attended the u of texas at austin which is part of the U of texas system . see was a test pilot . he died on 1966-02-28 in st. louis . \\ 
  \hline
\end{tabular}
\caption{\label{table:webnlg} A sample in the original WebNLG dataset (above) and the delexicalized dataset (below). In this case, referring expression ``He'', which refers to \emph{Elliot\_See}, is first delexicalized as elliot\_see and then extracted from the text.}
\end{table}


In the delexicalized version of WebNLG corpus, references to other discourse entities in the pre- and pos-contexts are represented by their one-word ID removing quotes and replacing white spaces with underscores. Since our model does not rely on any entity ID, we replaced each ID with its corresponding referring expression. 



\subsection{Profile Generation} 
We create a profile for each entity that appears in the corpus as a form of external knowledge for REG. The profile consists of the first three sentences of the entity's Wikipedia page, which is retrieved automatically using the entity's Wikipedia ID and the Python wikipedia 1.4.0 library. We choose the first three sentences of the Wikipedia page since they usually contain the most important information about an entity such as its name, country, or occupation.

We tokenized the crawled Wikipedia sentences and removed all phonetic symbols as well as other special characters. For the Wikipedia IDs that are ambiguous (e.g., New York may refer to New York City or New York (State)), we manually checked the discourse context to determine which entry was appropriate. For the entities that are constant numbers or dates which have no meaningful profiles, we make the profile content the same as the entity ID but replaced each underscore in the ID with white spaces. We generated a total of 1,501 profiles, one per entity in the dataset.

\subsection{Setups} \label{subsection:data}
In the current way that WebNLG is set up for REG, entities in the training set and test set are overlapped, which makes it impossible to examine the effect that previously unseen entities have on REG performance. To address this limitation, we split the dataset in three different ways as follows:


\paragraph{Original.} The original split of data from \citet{ferreira2018neuralreg}. There are $63{,}061$, $7{,}097$ and $8{,}743$ referring expressions in the training, development and test set respectively. In this dataset, all entities in the development and test set also appear in the training set. 

\paragraph{Entity-separated.} We separated the entities in the training, development and test set by splitting the delexicalized WebNLG dataset by entity IDs. The ratio of the number of entities in the training set, validation set, and test set is 8:1:1. As a result, there are $63{,}840$, $7{,}978$ and $7{,}083$ referring expressions of different entities in each dataset.

\paragraph{Random.} We randomly split all referring expressions in WebNLG with ratio 8:1:1. The final training, development and test set contains $63{,}121$, $7{,}890$ and $7{,}890$ referring expressions respectively. In this dataset, some entities in the test set appear in training, but not all. We also test our models on this dataset since we believe it better reflects a realistic setting in which a REG model might be applied.

\begin{figure*}[htbp]
\centering
\includegraphics[scale=0.48]{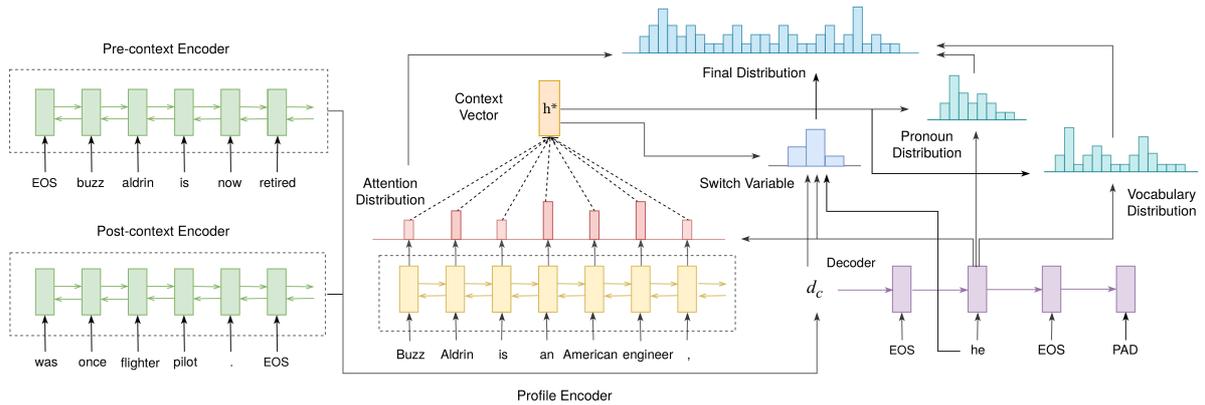}
\caption{Our \modelname{} model. The model uses two bi-directional LSTMs (green) to encode the pre- and post-context and one bi-directional LSTM (yellow) to encode the profile. At the decoding step, we calculate three distributions: attention distribution, pronoun distribution and vocabulary distribution. These three distributions together with the switch variable are used to determine the final distribution. }
\label{picture:model}
\end{figure*}

\section{Model}
In this section, we present our model, \modelname{} (Figure~\ref{picture:model}). In our task setup, the input of the model consists of pre- and post-context texts of lengths $H$ and $N$, and an entity profile of length $L$. The output is a sequence of tokens that make up the referring expression. We use three bidirectional LSTM \citep{hochreiter1997long} encoders to encode the pre- and post-context and the profile, and a unidirectional LSTM decoder to generate the output at inference time. We also apply an attention mechanism and a switch variable to calculate the final output distribution. We now introduce each component in more detail.

\subsection{Context Encoders}
We use two bi-directional LSTMs to encode the pre- and post-contexts, which are the entire sequences of tokens before and after the referring expression in the discourse context. The pre-context, a sequence of tokens $ \{w^{(pre)}_1, ..., w^{(pre)}_H\} $, is first mapped to a distributed vector space through an embedding layer, resulting in a sequence of vectors $ \{x^{(pre)}_1, ..., x^{(pre)}_H\} $. Then, the word embeddings are given to a bidirectional LSTM. We call the forward and backward hidden-state vectors at time $ t $ $ \overrightarrow{h}^{(pre)}_t $ and $ \overleftarrow{h}^{(pre)}_t $ respectively. We extract the two hidden-state vectors at the last step time $ H $,  and concatenate them to form the final vector representation $ h^{(pre)}_H = [\overrightarrow{h}^{(pre)}_H; \overleftarrow{h}^{(pre)}_H] $, which summarizes the information in the entire pre-context. The same process is repeated for the post-context using another bi-directional LSTM. This results in $ h^{(post)}_N = [\overrightarrow{h}^{(post)}_N; \overleftarrow{h}^{(post)}_N] $. Finally, we concatenate the pre- and post-context vector representations and pass them through a hidden layer:
\begin{equation}
    d_c = \tanh(W_d[h^{(pre)}_H; h^{(post)}_N]) 
\end{equation}
where $ W_d $ is a weight matrix mapping the concatenated hidden state vectors into a joint vector representation $ d_c $, which is used as the initial state of the decoder.


\subsection{Profile Encoder}
The profile encoder is another bi-directional LSTM that receives a series of profile words $ \{w_1, ..., w_L\} $ as input. All words in the profile are first converted to lowercase and mapped to a sequence of word embeddings $ \{x_1, ..., x_L\} $ through the same embedding layer as for the pre- and post-contexts. Then the embedded sequence are passed through the bi-directional LSTM, resulting in forward and backward hidden states $ \{\overrightarrow{h}_1, ..., \overrightarrow{h}_L\} $ and $ \{\overleftarrow{h}_1, ..., \overleftarrow{h}_L\} $. At each word position $ t $, the forward and backward representations are concatenated as the final representation $ h_t = [\overrightarrow{h}_t;\overleftarrow{h}_t] $. 


We also calculate a character-level embedding for each word using another bidirectional LSTM encoder. The last hidden-state vector of the forward and backward LSTMs are concatenated and given to a separate non-linear layer to form an output vector. This output vector is concatenated with the word-level embedding as our final word representation.



\subsection{Attention-based Decoder}
Our referring expression decoder is a unidirectional LSTM. At each decoding step $ t $, the decoder receives the word embedding of the previous word and generates a decoder hidden state $ s_t $. Then, we calculate the attention distribution $ a_t $ over the entity profile using the attention mechanism introduced by \citet{bahdanau2014neural}:
\begin{align}
    e^t_i &= v^{T} \tanh(W_h h_i + W_s s_t + b_{attn}) \\
    a^t &= {\rm softmax}(e^t),
\end{align}
where $ h_i $ is the sequence of profile encoder hidden states, $ W_h, W_s, v^T $ and $ b_{attn} $ are all trainable parameters. The attention distribution, $ a^t $, determines the degree of contribution of each token in the profile at expression generation step $ t $. A final summary $ h^*_t $ is calculated by summing the profile encoder states $ h_i $ weighted by the attention probabilities $ a^t $:
\begin{equation}
    h^*_t = \sum_i a_i^t h_i
\end{equation}
Then,  $ h^*_t $ is concatenated with the decoder hidden state $ s_t $ and passed through a linear layer, producing the vocabulary distribution $P_{voc}$:
\begin{equation}
    P_{voc} = {\rm softmax}(V[s_t; h^*_t] + b)
\end{equation}
where $ V $ and $ b $ are both trainable parameters in the model. $ P_{voc} $ contains the generation probability of each word in the vocabulary. We will actually modify this distribution to account for out-of-vocabulary items and pronouns by introducing a copying mechanism, which we will discuss in the next section. This results in the final generation probability, $ P(w_t^*) $ of the target (gold-standard) word at time $t$.

During training, we take the negative log likelihood of the target word in the gold standard, $\hat{w}_t^*$, as the loss of the model at step $ t $. The overall loss is the sum of the loss at each time step:
\begin{equation} \label{equation:loss}
    {\rm loss} = \frac{1}{T}\sum^T_{t=0} -\log P(\hat{w}_t^*)
\end{equation}

At test time, the word with the highest generation probability will be the output at the current time step and the input to the next step. 

\subsection{Switch Mechanism} \label{subsection:switch}
In \modelname{}, we apply a switch mechanism that allows the model to generate different referential forms, inspired by the pointer-generator network of \citet{see-etal-2017-get}. In particular, our model can choose to: i) \textbf{copy} a word from the profile, which is especially useful for named entities with rare names which are not part of a fixed vocabulary, ii) generate a \textbf{pronoun}, or iii) \textbf{generate} from a fixed vocabulary. Although pronouns are part of our fixed vocabulary, we choose to distinguish them as a separate category to model, due to their importance and frequency in the REG task.

We define a switch variable $\Sigma$ which can take on one of three values, \textsc{Copy}, \textsc{Pro}, and \textsc{Gen}, corresponding to the three decoder actions. Let the associated probabilities be $\sigma_{\textsc{copy}}$, $\sigma_{\textsc{pro}}$, and $\sigma_{\textsc{gen}}$. 
After computing $ h^*_t $ and vocabulary distribution $ P_{voc} $ at time step $ t $, we compute these probabilities using the summary vector $ h^*_t $, the encoder final state $ d_c $, the decoder hidden state $ s_t $ and the input $ x_t $:
\begin{equation} 
\begin{split}
    & [\sigma_{\textsc{copy}}, \sigma_{\textsc{pro}}, \sigma_{\textsc{gen}}] = {\rm softmax}( \\
    & \quad H_{h^*} h^*_t + H_d d_c + H_s s_t + H_x x_t + b_{s}),
\end{split}
\end{equation}
where $ H_{h^*}, H_d, H_s, H_x $ and $ b_{s} $ are learnable parameters with output dimension of three. The final generation probability of the target word $ w^*_t $ is then computed as follows:
        
\begin{spacing}{1.15}
\begin{algorithmic}
  \IF{$ w^*_t $ is a pronoun}
    \STATE $ P(w^*_t)= \sigma_{\textsc{pro}}P_{voc}(w^*_t) + \sigma_{\textsc{copy}}\sum\limits_{i:w_i=w^*_t}a_i^t $
  \ELSIF{$ w^*_t $ is a vocabulary word}
    \STATE $ P(w^*_t)= \sigma_{\textsc{gen}}P_{voc}(w^*_t) + \sigma_{\textsc{copy}}\sum\limits_{i:w_i=w^*_t}a_i^t $
  \ELSE
    \STATE $ P(w^*_t) = \sigma_{\textsc{copy}}\sum\limits_{i:w_i=w^*_t}a_i^t $
  \ENDIF
\end{algorithmic}
\end{spacing}
In the formula, $ a^t $ is the attention distribution computed in the previous section. When $ w^*_t $ is a pronoun or vocabulary word, the probability $ P(w^*_t) $ consists of two terms since the word in the vocabulary could also be in the profile. $ P(w^*_t) $ will be used in the Equation~\ref{equation:loss}. 

One important difference between our model and the pointer-generator network is that the pointer-generator model only computes one \emph{generation probability} $ p_{gen} \in [0, 1] $, which is used to decide whether generate a word or copy a token from the source document. In our model, however, we compute a probability distribution $ S_{switch} $ over three values. 
Our intention is that the model can better distinguish between different forms of referring expressions, especially between pronouns and demonstrative or definite description forms.

\section{Experiments}
We evaluate our model and the baselines on three different data splits mentioned in Section~\ref{subsection:data}: \textbf{Original}, \textbf{Entity-separated} and \textbf{Random}, where the entities are overlapped, all separated and randomly mixed in training, validation and test set.

\subsection{Baselines} \label{subsection:baseline}
We compared our model against three baselines: \emph{OnlyName}, \emph{Ferreira} and \emph{NeuralREG}. 
In our first \emph{OnlyName} baseline, we use the name of the entity as the final realization regardless of the context. This baseline is inspired by the fact that 71.4\% of the referring expressions in WebNLG are proper names. Given the Wikipedia profile of an entity, we obtain the name of the entity by extracting the first few capitalized words in that profile. If the name cannot be found in this way, we will use the first word of the profile (except articles).

Our second baseline makes use of the Naive Bayes model proposed by \citet{ferreira2016towards}, dubbed \emph{Ferreira}. This model takes features including syntactic position, referential status (initial or a subsequent mention at the level of text and sentence) and recency (distance between references) as input. 
The original model is only used to determine the form of the referring expression. We adapt the model by adding a content generation component. Once the form of the referent is decided, the content of the referential expression is decided as follows: we use the \emph{OnlyName} model to get the name of the entity and find the most frequent pronoun expressions in the profile as the content for pronoun form (using ``it'' if no pronoun expressions can be found). We use the gold-standard content for the demonstrative and description categories from the WebNLG dataset, which provides an upper bound on the potential performance of this system.

Our third baseline is the NeuralREG model proposed by \citet{ferreira2018neuralreg}. NeuralREG is a sequence-to-sequence neural network model with an attention mechanism. All words and entities in NeuralREG are represented as embeddings which are randomly initialized and learned during training. We tested the best-performing CAtt version of NeuralREG.

\subsection{Metrics}
Following prior work, we calculate the overall accuracy and the String Edit Distance (SED) between the generated and the reference expressions. The overall accuracy is calculated by comparing the generated string with the real referring expression. We post-processed the output to ignore differences in capitalization, accent and certain non-alphabetic characters (e.g., ``Desteapta-te'' vs. ``De\c steapt\u a-te'', and ``Tranmere Rovers FC'' vs. ``Tranmere Rovers F.C.'').

We also evaluate the performance of our model with regard to a particular referential form. We computed name accuracy, pronoun accuracy, precision, recall and F1-score. The name and pronoun accuracy are computed by comparing names and pronouns in the test set with the generated referring expressions. Pronoun classification evaluation takes all generated referring expressions as two forms: pronoun or non-pronoun and do not consider the actual content.


\subsection{Experiment Setting}
For all experiments, our model has 100-dimensional hidden states for the encoder and decoder LSTMs and 50 dimensional for the character LSTM. The word and character embeddings are both set to 100-dimensional. We initialize the word embeddings using pre-trained GloVe vectors \cite{pennington2014glove} and all character embeddings are randomly initialized. The model is trained for up to 35 epochs, or until the performance on the development set does not improve for 5 epochs. We set the learning rate to 0.0037 and the batch size to 64. We apply gradient clipping with a maximum gradient norm of 5. We adopt dropout training with a dropout rate of 0.5. At test time, all referring expressions are generated using greedy decoding. These settings were selected by tuning on the development set.

\section{Results}

\begin{table*}[t!]
\renewcommand{\arraystretch}{1.14}
\centering
\begin{tabular}{c|c|c|c|c|c|ccc}
  \toprule
  \multirow{2}{*}{\bf Dataset} & \multirow{2}{*}{\bf Models} & \multirow{2}{*}{\bf SED} & \multirow{2}{*}{\bf Total Acc.} & \multirow{2}{*}{\bf \makecell{Name Acc.} } & \multirow{2}{*}{\bf \makecell{Pronoun \\ Acc.} } & \multicolumn{3}{c}{\bf Pronoun Classification} \\
  ~ & ~ & ~ & ~ & ~ & ~ & Prec. & Rec. & F1  \\
  \hhline{=========} 
  \multirow{4}{*}{\bf Original} & OnlyName & 5.57 & 53.47\% & 66.22\% & 0.00\% & 0.00 & 0.00 & 0.00 \\
  ~ & Ferreira & 4.56 & 56.50\% & 65.11\% & 63.58\% & 0.60 & 0.96 & 0.74 \\
  ~ & NeuralREG & 2.33 & 74.13\% & \bf 81.64\% & 74.92\% & 0.77 & 0.78 & 0.78 \\ \cline{2-9}
  ~ & \modelname{} & \bf 2.22 & \bf 74.67\% & 80.51\% & \bf 87.17\% & 0.72 & 0.89 & \bf 0.80 \\ 
  \hhline{=========}
  \multirow{4}{*}{\bf \makecell{Entity- \\ separated}} & OnlyName & 6.28 & 54.71\% & 61.44\% & 0.00\% & 0.00 & 0.00 & 0.00 \\
  ~ & Ferreira & 5.91 & 55.67\% & 60.11\% & 64.25\% & 0.45 & 0.97 & 0.61 \\
  ~ & NeuralREG & -- & -- & -- & -- & -- & -- & -- \\ \cline{2-9}
  ~ & \modelname{} & \bf 5.17 & \bf 60.60\% & \bf 67.20\% & \bf 75.57\% &  0.53 & 0.80 & \bf 0.64 \\
  \hhline{=========}
  \multirow{4}{*}{\bf Random} & OnlyName & 5.59 & 53.21\% & 67.02\% & 0.00\% & 0.00 & 0.00 & 0.00 \\
  ~ & Ferreira & 4.72 & 55.63\% & 65.59\% & 61.37\% & 0.58 & 0.95 & 0.72 \\
  ~ & NeuralREG & 2.61 & 71.62\% & 77.49\% & 75.03\% & 0.61 & 0.89 & 0.71 \\ \cline{2-9} 
  ~ & \modelname{} & \bf 2.11 & \bf 75.33\% & \bf 81.85\% & \bf 83.89\% & 0.76 & 0.85 & \bf 0.80 \\
  \bottomrule
\end{tabular}
\caption{\label{table:all-results} REG performance on the original, entity separated and random dataset. \modelname{} is the model proposed in this work. Note that NeuralREG cannot be applied to the entity-separated dataset.}
\end{table*}

\subsection{Automatic Evaluation}




As shown in Table~\ref{table:all-results} \modelname{} outperforms the three baselines in all experiments. 
Our model in particular excels at the measures of pronoun accuracy and pronoun classification. 

On the \textbf{Original} dataset, our model is slightly better than NeuralREG in total accuracy and SED. This result indicates that our model is as good as or better than NeuralREG model when dealing with previously seen entities. On pronoun classification, \emph{OnlyName} model has a performance of zero since it cannot generate any pronouns. 

\textbf{Entity-separated} demonstrates our model's ability to handle new entities, compared to the baselines. There is a clear drop on most metrics compared to the original dataset, but this is expected since the test set consists of unseen entities. 
As previously explained, the NeuralREG model cannot be applied to this dataset without extensive modifications. On \textbf{Random}, \modelname{} outperforms NeuralREG. For the unseen entities, NeuralREG generates a random string, since the embeddings of unseen entities are not updated by training. Our model, however, is capable of producing accurate referring expressions for both previously mentioned and new entities.

\begin{table*}[t!]
\small
\renewcommand{\arraystretch}{1.35}
\centering
\begin{tabular}{p{15.5cm}}
  \toprule
  \emph{Original}: {\bf Acharya Institute of Technology} is affiliated with Visvesvaraya Technological University which is in Belgium . {\bf The institute itself} is in India 's Karnataka state and {\bf its} full address is In Soldevanahalli , Acharya Dr. Sarvapalli Radhakrishnan Road , Hessarghatta Main Road , Bangalore -- 560090 . {\bf It} was created in 2000 and {\bf its} director is Dr. G. P. Prabhukumar . \\ 
  \hline
  \emph{OnlyName}: {\bf Acharya Institute of Technology} is affiliated with Visvesvaraya Technological University which is in Belgium . {\bf Acharya Institute of Technology} is in India 's Karnataka state and {\bf Acharya Institute of Technology} full address is In Soldevanahalli , Acharya Dr. Sarvapalli Radhakrishnan Road , Hessarghatta Main Road , Bangalore -- 560090 . {\bf Acharya Institute of Technology} was created in 2000 and {\bf Acharya Institute of Technology} director is Dr. G. P. Prabhukumar . \\ 
  \hline
  \emph{Ferreira}: {\bf Acharya Institute of Technology} is affiliated with Visvesvaraya Technological University which is in Belgium . {\bf It} is in India 's Karnataka state and {\bf its} full address is In Soldevanahalli , Acharya Dr. Sarvapalli Radhakrishnan Road , Hessarghatta Main Road , Bangalore - 560090 . {\bf It} was created in 2000 and {\bf it} director is Dr. G. P. Prabhukumar . \\ 
  \hline
  \emph{NeuralREG}: {\bf Acharya institute of technology} is affiliated with Visvesvaraya Technological University which is in Belgium . {\bf It} is in India 's Karnataka state and {\bf its} full address is In Soldevanahalli , Acharya Dr. Sarvapalli Radhakrishnan Road , Hessarghatta Main Road , Bangalore -- 560090 . {\bf It 's} was created in 2000 and {\bf its} director is Dr. G. P. Prabhukumar . \\ 
  \hline
  \emph{\modelname{}}: {\bf The Acharya Institute of Technology} is affiliated with Visvesvaraya Technological University which is in Belgium . {\bf The institute} is in India 's Karnataka state and {\bf its} full address is In Soldevanahalli , Acharya Dr. Sarvapalli Radhakrishnan Road , Hessarghatta Main Road , Bangalore -- 560090 . {\bf It} was created in 2000 and {\bf its} director is Dr. G. P. Prabhukumar . \\
  \hhline{=}
  \emph{(Entity profile)} Acharya Institute of Technology , or AIT , is a private co-educational engineering and management college in Bengaluru , India , affiliated with the Visvesvaraya Technological University and accredited by the National Board of Accreditation . Established in 2000 , it offers eleven undergraduate courses and eight postgraduate courses . The college has links and collaborations with various industries and universities across the world . \\ 
  \bottomrule
\end{tabular}
\caption{\label{table:example} Example of original text and generated text of each model.}
\end{table*}

Table~\ref{table:example} shows an example of the generated text of each model. The entity in this example is \emph{Acharya Institute of Technology}. Note that only the bold text is generated by the models. The output of the \emph{OnlyName} model is clearly the least readable one.

\subsection{Seen Entities vs. Unseen Entities}
\begin{table}[t!]
\renewcommand{\arraystretch}{1.05}
\begin{center}
\begin{tabular}{c|c|c|c}
\toprule ~ & \bf Type & \bf Acc. & \bf Support \\ 
\hline
\multirow{5}{*}{\bf Seen} & demonstrative & 0.00\% & 22 \\
~ & description & 48.72\% & 862 \\
~ & name & 79.11\% & 2547 \\
~ & pronoun & 90.00\% & 160 \\
\cline{2-4}
~ & total & 71.82\% & 3591 \\
\hline
\multirow{5}{*}{\bf Unseen} & demonstrative & 0.00\% & 3 \\
~ & description & 20.54\% & 409 \\
~ & name & 74.74\% & 2423 \\
~ & pronoun & 88.33\% & 120 \\
\cline{2-4}
~ & total & 67.72\% & 2955 \\
\bottomrule
\end{tabular}
\end{center}
\caption{\label{analysis-table} Evaluation for seen and unseen entities. }
\end{table}

In the evaluation, we also distinguished the results for seen and unseen entities. We trained the model on a training set contains 64,353 referring expressions and evaluate the model on a test set with 3,591 referring expressions related to seen entities and 2,955 expressions related to unseen entities. Table~\ref{analysis-table} shows the evaluation results. From Table~\ref{analysis-table}, it is easy to see that the model performs better when generating referring for seen entities. Among four referring expression types, the accuracy of description type drops dramatically for unseen entities, from 48.72\% to 20.54\%. This is probably due to the fact that compared with name and pronoun, description type is often hard to identify and more flexible. For instance, one of the gold-standard descriptions in the test set is \emph{the comic character , amazing-man}. The model's generation for this referring expression is \emph{amazing-man}. 

Also notice that the accuracy of demonstrative type is zero for both seen and unseen entities. We think this is because the dataset is unbalanced and there are too few demonstrative samples in the training set. The model did not have enough samples to learn to identify demonstratives. 

\subsection{Human Evaluation}
\paragraph{Method.}
In addition to automatic evaluation, we also performed a human evaluation, comparing our model to the original text and to the three baseline models. We concatenated the pre- and post-contexts with the generated referring expression to form an overall passage. We showed participants the output of our model and a comparison model, and asked them to make a pairwise preference judgment between the two passages.

Specifically, we randomly selected 100 samples from the \textbf{Random} test set. We presented the two passages in random order, asking participants to consider the readability, grammaticality and fluency of texts. We then asked them to select the passage that they prefer, or to say they are equally good. We recruited 20 participants, each evaluating 20 samples.  


\begin{figure}[t!]
\centering
\includegraphics[scale=0.53]{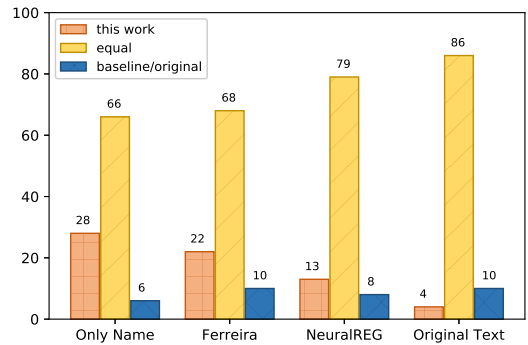}
\caption{Human evaluation results. Comparison of our model with three baselines and original text.}
\label{picture:he}
\end{figure}

\paragraph{Results.}
Figure~\ref{picture:he} shows the human evaluation results. The most common choice made by participants is ``equal''. This is because many generated referring expressions by the models are actually identical. When our model was compared to the original text, 86 of the 100 generated texts are identical or very similar to the original text. This demonstrates our model's ability to generate high quality referring expressions. Also, we can see from the chart that our model tends to be preferred over the three baseline models. 

\begin{figure}[t!]
\centering
\includegraphics[scale=0.53]{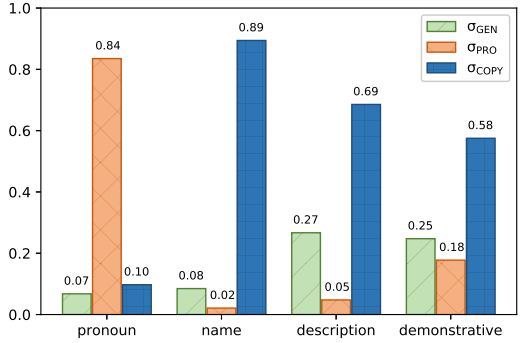}
\caption{The average probabilities of the switch variable values for each referential form on the \textbf{Random} test set.}
\label{picture:chart}
\end{figure}

\subsection{Analyzing Switch Variable Behaviour}
We now examine the behaviour of the switch variable, $\Sigma$. As mentioned in Section~\ref{subsection:switch}, each word is generated with a probability distribution  ($ \sigma_{\textsc{copy}} $, $ \sigma_{\textsc{pro}} $ and $ \sigma_{\textsc{gen}} $). We expect this distribution to vary based on the form of the referring expression. For instance, when a proper name expected, $ \sigma_{\textsc{copy}} $ should be relatively high if the model copies words from the entity profile. 

Table~\ref{table:sample} shows three examples of generated referring expressions with the values of the switch variable. We put the gold-standard referring expression, pre- and post-context together in Context. The words in yellow, green and orange highlight are copied words, vocabulary words and generated pronouns respectively and each word is followed by the corresponding switch value. These examples show that our model is able to switch between the three actions to generate a coherent, fluent referring expression. We further show the average value of the switch variable on the \textbf{Random} test set separated by the gold-standard referential form (Figure~\ref{picture:chart}).



\definecolor{MyGreen}{rgb}{0.76,0.88,0.71}
\definecolor{MyBlue}{rgb}{0.18,0.33,0.80}
\definecolor{MyOrange}{rgb}{0.97,0.80,0.68}

\begin{table}[t!]
\small
\renewcommand{\arraystretch}{1.2}
\setlength\tabcolsep{3.5pt}
\setlength{\belowcaptionskip}{-0.46cm}
\centering
\begin{tabular}{|p{7.3cm}|}
  \hline
  {\bf Context}: The 3rd runway at \textit{Ardmore Airport ( New Zealand )} is made of Poaceae . (...) \\ 
  {\bf Entity Profile}: Ardmore Airport is an airport 3 NM southeast of Manurewa in Auckland , New Zealand . (...) \\
  {\bf Realization}: 
  \hl{$\text{Ardmore}_{\text{(COPY 0.968)}}$} \hl{$\text{Airport}_{\text{(COPY 0.971)}}$} \hlc[MyGreen]{$\text{in}_{\text{(GEN 0.908)}}$} \hl{$\text{New}_{\text{(COPY 0.969)}}$} \hl{$\text{Zealand}_{\text{(COPY 0.947)}}$} \\ 
  \hline 
  {\bf Context}: Curitiba is part of Parana State in the South Region , Brazil and is served by Afonso Pena International airport . It is led by \textit{the Democratic Labour Party} . \\ 
  {\bf Entity Profile}: The Democratic Labour Party is a social democratic political party in Brazil . (...) \\
  {\bf Realization}: 
  \hlc[MyGreen]{$\text{the}_{\text{(GEN 0.895)}}$} \hl{$\text{Democratic}_{\text{(COPY 0.938)}}$} \hl{$\text{Labour}_{\text{(COPY 0.987)}}$} \hl{$\text{Party}_{\text{(COPY 0.966)}}$} \\
  \hline
  {\bf Context}: Elliot See was born on July 23 , 1927 in Dallas . \textit{He} attended the U of Texas at Austin which is part of the U of Texas system . (...) \\ 
  {\bf Entity Profile}: Elliot See was an American engineer , naval aviator , test pilot , and NASA astronaut . He was selected for NASA 's second group of astronauts in 1962 . \\
  {\bf Realization}: \hlc[MyOrange]{$\text{he}_{\text{(PRO 0.987)}}$} \\ 
  \hline
\end{tabular}
\caption{\label{table:sample} An example of how the switch variable works in referring expression generation.}
\end{table}



\section{Conclusions}
In this work, we have proposed new test setups on WebNLG dataset that evaluate the performance of REG models when dealing with previously unseen entities. We also introduced \modelname{}, an end-to-end deep neural network model for REG. Unlike previous REG models that only model the local context, \modelname{} incorporates both the context information and an entity profile that allows it to generate references for unseen entities. Our results show that \modelname{} can accurately generate referring expressions for both new and previously mentioned entities, outperforming three strong baselines. One future direction is to model larger contexts with this approach. It would also be interesting to integrate our model into an existing language generation system (e.g., as part of an abstractive summarization system).

\section*{Acknowledgments}
We would like to thank the reviewers for their valuable comments. This research was supported by the Mitacs Globalink Program and the Canada CIFAR AI Chair program.


\bibliography{emnlp-ijcnlp-2019}
\bibliographystyle{acl_natbib}

\appendix

\end{document}